\documentclass[]{spie}  

\usepackage{amsmath,amsfonts,amssymb}
\usepackage{graphicx}
\usepackage[colorlinks=true, allcolors=blue]{hyperref}
\usepackage[table]{xcolor}
\usepackage{tabularx}
\usepackage{makecell}
\usepackage{tikz}
\usepackage{fontawesome}
\newcommand*\Circle[1]{\protect\tikz[baseline=(char.base)]{
            \protect\node[shape=circle,draw,inner sep=0.5pt] (char) {\footnotesize{#1}};}}

\usepackage[textsize=small]{todonotes} 
\setuptodonotes{inline}

\title{Beyond Benchmarks: Using VLMs to Reveal Systematic Classification Failures Under Real World Conditions}

\author[a*]{Dieuwertje Alblas}
\author[a*]{Alma M. Liezenga}
\author[a]{Jan Erik van Woerden}
\author[a]{Fedor Taggenbrock}
\author[a]{Dalia Aljawaheri}
\author[a]{Klamer Schutte}
\affil[a]{Intelligent Imaging, Defense, Safety \& Security, TNO, The Hague, The Netherlands}

\authorinfo{Further author information: (Send correspondence to D.A.)\\D.A.: E-mail: dieuwertje.alblas@tno.nl\\
* These authors contributed equally to this work}

\begin{document} 

\maketitle

\begin{abstract}
Verification and validation (V\&V) of classification models is crucial to enable a wide range of sensor processing applications. Currently, the V\&V process relies on time-consuming manual inspection of erroneous samples to find meaningful patterns. This work explores the use of Vision Language Models (VLMs) to speed up this laborious process. VLMs are trained to embed images into a semantically meaningful vector representation, from which human-interpretable systematic errors can be distilled. Deploying such VLM-based methods in a defence context introduces two major challenges: (1) the defence domain is underrepresented in the training data of VLMs, and (2) surroundings and context are less diverse than for other domains. This study provides an initial assessment of the suitability of VLM-based methods for V\&V of defence applications. We propose a VLM-based error slice detection (ESD) method that independently groups and labels systematic errors made by a classification model. We demonstrate that this method is able to identify operationally-relevant artificially added perturbations in a non-military dataset. In a military context, our method clusters and describes images based on their surroundings, but also exhibits overlap between cluster descriptions. We further investigate the difference in embedding variation between our military and non-military dataset, which remains a topic of interest. Although the results do not yet warrant fully automated V\&V through VLM-based ESD, they show that VLMs could be used to accelerate V\&V processes in the future. 
\end{abstract}


\section{INTRODUCTION}\label{sec:intro} 
Modern military vehicles are equipped with an increasing number of sensors to enhance situational awareness, such as cameras for target detection and recognition. This growth leads to high-volume sensor data streams, necessitating automatic analysis of sensor measurements \cite{AIFactsheetDef}. Artificial intelligence (AI), particularly discriminative AI, has shown promising results in, e.g. small object detection \cite{van2024toward}, fine-grained military vehicle classification \cite{van2025occlusion} and classification of vessels \cite{den2021vessel}. To facilitate safe and reliable deployment in operational scenarios, verification and validation (V\&V) of AI models is crucial but poses a significant challenge \cite{paardekooper2024toward, fokkinga2024validation}. In this work, we focus on V\&V of discriminative AI methods operating on visual data. These vision models may exhibit degraded performance under conditions such as adverse weather, occlusions, camouflage, or smoke, which frequently occur in operational environments \cite{mirza2021robustness, nijskens2023predicting}. Being aware of these shortcomings prior to deployment and mitigating their impact is important for the safe and responsible deployment of AI models in operational contexts.

AI models are black-box by design and learn non-linear mappings between input and output from a large set of training examples. The black-box nature of these models results in unclear decision boundaries. The essentially unlimited space of possible input data specific to vision further complicates comprehensive testing of AI models. Moreover, model strongly depends on the training data; typically AI models do not generalize well to out-of-distribution samples \cite{liu2021towards}. When using third party models, there is potentially limited knowledge about the data that was used to train the AI model, which makes identifying model vulnerabilities difficult. Together, these aspects make systematic identification of failure modes a complex process, limiting the confidence in a model's behaviour under previously unseen but operationally relevant conditions.

Traditional V\&V methods such as failure modes and effect analysis (FMEA) \cite{bluvband2009failure} are commonly applied to rule-based systems, where system behaviour is explicitly defined. However, for AI models with complex and implicit decision boundaries, some parts of FMEA should be reconsidered as a full sweep of potential inputs is not feasible. Consequently, AI models are primarily evaluated based on their observed behaviour on a finite set of test samples. For V\&V purposes, it is crucial to understand \textit{why} a model fails in specific test cases. For example, which environmental factors, visual attributes or conditions lead to incorrect predictions. This knowledge can then be used to mitigate vulnerabilities through, e.g. design choices or model fine-tuning.

One commonly used approach to gain insight into model vulnerabilities is explainable AI. For vision models, saliency-based methods such as Grad-CAM \cite{selvaraju2020grad} are a well-established technique. These methods generate heatmaps that indicate which image regions are most strongly associated with a model's prediction. While saliency maps can provide qualitative insights into individual model decisions, extracting actionable knowledge from them is labor-intensive. In practice, an operator should manually inspect large numbers of images and interpret the highlighted regions to categorize failure cases on image semantics. This manual step limits the scalability for systematic V\&V of AI models.

Until recently, semantic reasoning over image data was restricted to human interpretation. The emergence of contrastive vision-language models (VLMs) fundamentally changes this restriction. Contrastive VLMs embed images and text into a shared latent space, which is structured by semantic concepts \cite{radford2021learning}. As a result, images containing similar semantic attributes are mapped close together in this latent space, and distance measurements can be interpreted as a notion of semantic similarity \cite{papadimitriou2025interpreting}. This 
opens up new avenues for automatic interpretation of image semantics. Generative VLMs, on the other hand, take image and/or text input and generate textual (or visual) responses based on them \cite{raja2025advancing}, offering even more opportunities for automated semantic interpretation. 

Recent works have leveraged VLMs for automatic error slice discovery (ESD), i.e. the automatic discovery of systematic errors of AI models. We distinguish two types of ESD methods. First, \textit{slice-then-tag} methods that leverage the semantic structure of the embedding space provided by a contrastive VLM to detect error clusters, after which cluster descriptions are automatically derived using a generative VLM \cite{eyubogludomino,jain2022distilling,yenamandra2023facts}. Second, \textit{tag-then-slice} methods \cite{chen2025hibug2,chen2023hibug}, where this order is reversed: a generative VLM is used to label images based on a set of predetermined attributes and tags. In this work, we focus on the first type of methods.

Applying VLMs for automatic ESD in defence-related scenarios introduces two key challenges. First, the defence domain is underrepresented in VLM training data, which may result in inaccurate or unreliable image embeddings. Second, these ESD methods are often developed using academic datasets captured under ideal conditions and containing a wide variation of environments. In contrast, (operational) military data is subject to significantly lower data quality and covers a much smaller range of environments. In this work, we assess how VLM-based error slice discovery transfers from a well-represented domain to the military domain. To this end, we leverage a dataset with images of different dog breeds and multiple datasets with images of military vehicles. Moreover, we assess if VLM-based ESD can detect artificially introduced systematic perturbations in the data.

\section{MATERIALS \& METHODS}
In this work, we evaluate the value of VLMs in automatic error slice discovery for the purpose of V\&V of AI models in a defense context. VLMs have the capability to map images to a latent space, structured by semantics. We introduce a VLM-based automatic ESD method, inspired by the Domino method introduced in Eyuboglu et al \cite{eyubogludomino}.

\begin{figure}
    \centering
    \includegraphics{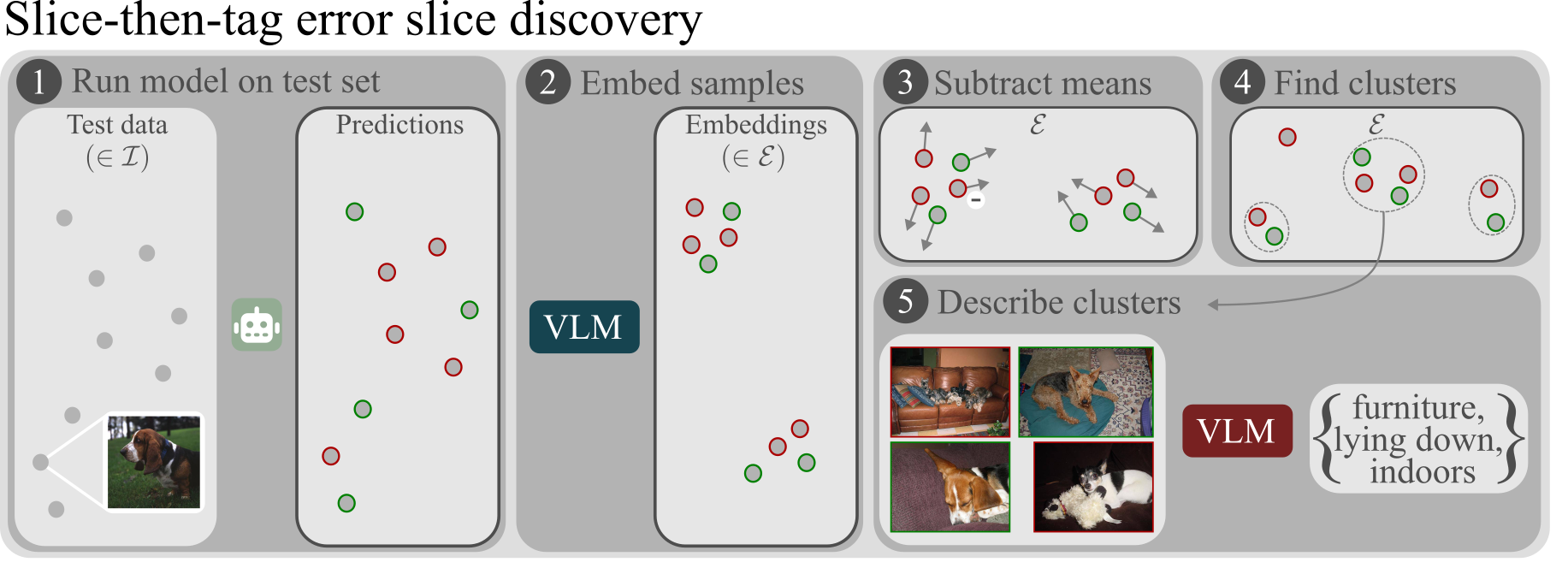}
    \caption{Overview of the VLM-based ESD method, consisting of five steps. \Circle{1} the AI model subject to V\&V is applied on a set of test images, resulting in correctly (green) and wrongly (red) predicted samples. \Circle{2} the test images are mapped to the joint text and image embedding space using the VLM. \Circle{3} to shift the embeddings' focus from the main object to its surroundings, the respective mean class embedding is subtracted from the sample embeddings. \Circle{4} clusters are retrieved from these residual embeddings. \Circle{5} images from a cluster are jointly fed to a generative VLM, which is prompted to describe the similarities between the images, in this case \textit{furniture}, \textit{lying down} and \textit{indoors}.}
    \label{fig:method}
\end{figure}

\subsection{VLM preliminaries}
The contrastive VLM that is central in our ESD method operates in three distinct spaces: (1) image space $\mathcal{I}$, (2) embedding space $\mathcal{E} \subset \mathbf{R}^d$, and (3) semantic space $\mathcal{S}$. The VLM learns mappings from both image space $\mathcal{I}$ and semantic inputs $\mathcal{S}$, e.g. text, to a shared $d$-dimensional embedding space $\mathcal{E}$. We consider a classification dataset $\mathcal{X} \subset \mathcal{I}$, consisting of $n$ images, each assigned to exactly one of $k$ classes. We denote the subset of images from $\mathcal{X}$ that belong to class $c \in {1, 2, 3,..., k}$ as $\{x_c\}$.

The spaces $\mathcal{I}, \mathcal{S}$, and $\mathcal{E}$ differ substantially in terms of geometric properties and interpretability. While elements from $\mathcal{I}$ and $\mathcal{S}$, i.e. images and text, are human-interpretable, the calculation and interpretation of (Euclidean) distances between elements in these spaces is ambiguous. In contrast, although individual points in the joint embedding space $\mathcal{E}$ are not human-interpretable, the geometry of this space is structured such that small distances reflect semantic similarity \cite{papadimitriou2025interpreting}.

Images typically comprise multiple objects and environmental elements, which are jointly encoded by a VLM into a single latent representation. For ESD, all these components are equally important. However, VLM embeddings tend to be dominated by the largest or most salient object in the image \cite{ruthardt2026steerable}. Recent work has shown that VLM latent spaces exhibit local linearity, with embeddings approximating linear combinations of semantic concepts \cite{papadimitriou2025interpreting,tewel2022zerocap}. This property enables the manipulation of embeddings to disentangle objects from contextual elements.

\subsection{VLM-based Error Slice Discovery}
Our VLM-based ESD method consists of five steps, as shown in \autoref{fig:method}. First, the AI model subject to V\&V is inferenced on a set of test images, resulting in a number of correctly and wrongly predicted samples. Second, the same test images are embedded using the VLM $g(x; \theta): \mathcal{I} \rightarrow \mathcal{E}$, resulting in embeddings $e_x \in \mathcal{E}$. Third, we subtract the class-level average embedding $e_c$ from the image embeddings ($e_x$) belonging to that corresponding class, with the idea of shifting the representation away from the main object and toward its surrounding environment. We calculate the mean class embeddings as $e_c = \frac{1}{|x_c|} \sum\limits_{x\in x_c} e_x$. Fourth, we perform dimensionality reduction on these residual embeddings and clustering in this lower dimensional space. Fifth, we yield a human-interpretable description of each cluster by leveraging a generative VLM. For each cluster, we create a collage containing multiple images that is fed to the VLM, which is prompted to describe the cluster using five words ranked from most important to least important\footnote{We deviate from Domino here, in which a textual description is selected from a corpus based on the similarity of its embedding to the weighted average embedding of the cluster.}. The full prompt is given in \autoref{app:prompt}.

These five steps result in automatically acquired, human-interpretable descriptions of semantically related test samples. Note that these clusters could contain both \textit{correctly} and \textit{wrongly} predicted samples, which gives insight into the model performance in a certain type of environment.

\begin{figure}[ht!]
    \centering
    \includegraphics[scale=0.9]{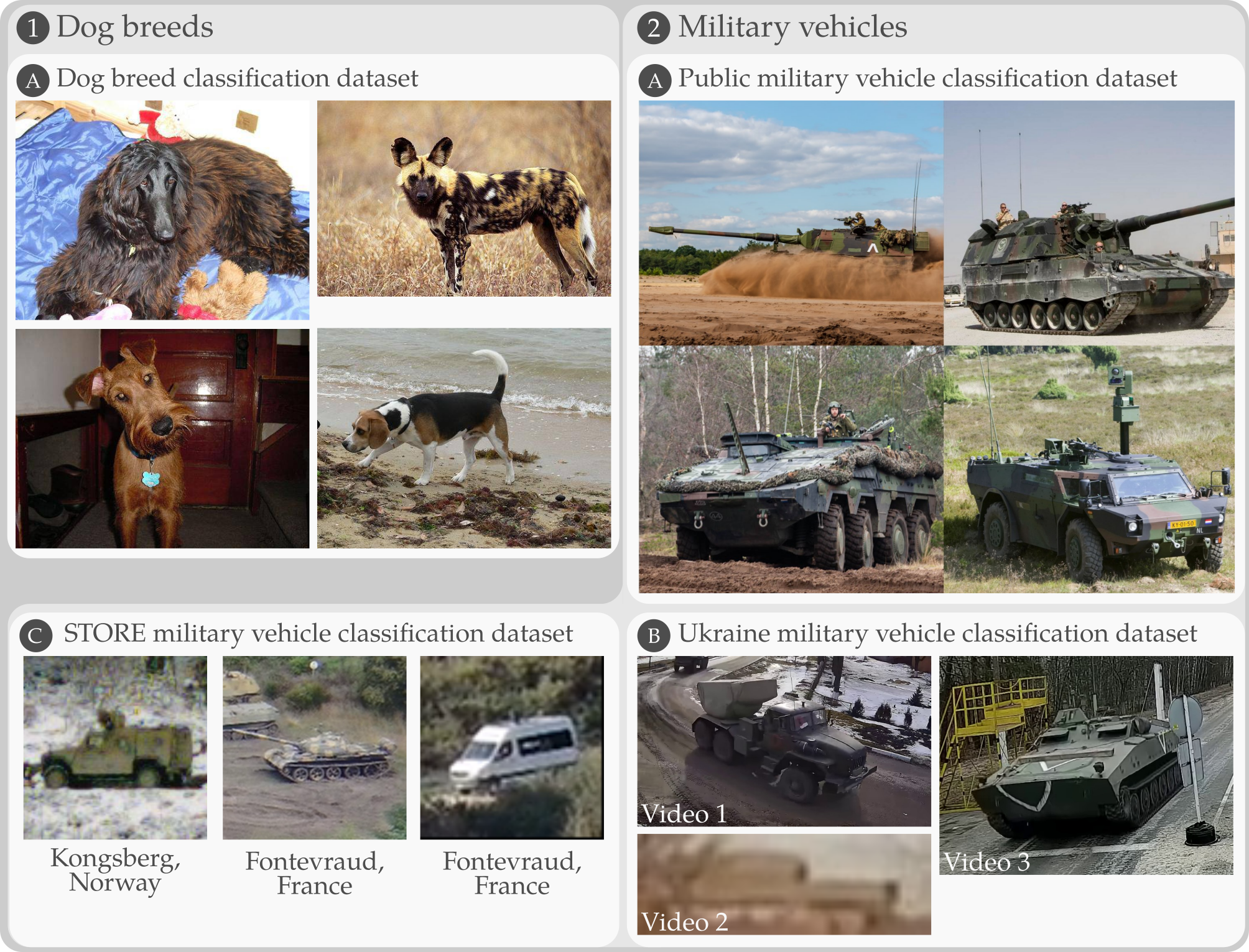}
    \caption{Overview of the datasets from two different domains covered in this work: \Circle{1} dog breeds and \Circle{2} military vehicles, consisting of one and three sources, respectively \protect\footnotemark. Datasets 1A and 2A are obtained from public resources, while datasets 2B and 2C are obtained from bounding boxes in video frames and represent more challenging conditions.}
    \label{fig:dataset}
\end{figure}

\subsection{Data}
In this work, we use four classification datasets from two domains: (1) dog breeds, and (2) military vehicles. \autoref{fig:dataset} shows an overview of these datasets, consisting of one dog breed dataset and three datasets containing military vehicles. We include these two domains as we expect a discrepancy in their representation in the training data of the VLM. Images of dogs in different surroundings are widely available on the internet, while images of military vehicles are less represented. Moreover, the surroundings of the images of dog breeds varies more than for the military vehicles. These two aspects could influence the quality of the VLM embeddings of these images, resulting in performance discrepancies of the ESD method between the two domains. We first describe the four datasets in more detail.

\textbf{1A: Dog breed classification dataset} \cite{kabilan_dogbreedclassification} consists of 8.040 web-based images of 93 different dog breeds. This dataset resembles the typical academic conditions, where image quality is high and the main objects are mostly centered. Nevertheless, this dataset contains a large variety in environments and small inter-class differences, making it useful for understanding strengths and weaknesses of the ESD method.

\footnotetext{2A: Top row: a Panzerhaubitze 2000 (left image credit: John van den Boogaart, defensiefotografie.nl; right image credit: Netherlands Ministry of Defence); bottom left: a Boxer (image credit: Martin Bos, defensiefotografie.nl); bottom right: a Fennek. 2B: Recorded during the STORE campaign in Kongsberg by Safran Electronics and Defense and in Fontevraud by FlySight S.r.l., respectively.}

\textbf{2A: Public military vehicles classification dataset} \cite{van2025occlusion, van2024visual} is an in-house dataset compiled from public sources and consists of 18 vehicle classes representing diverse military platforms: tanks (T-62, T-72, T-90, Leopard 2, M1A2), armored personnel carriers (BTR-80, BMP-1, Fuchs, Boxer, Patria), reconnaissance vehicles (BRDM-2, Fennek), self-propelled artillery (2S1, 2S3, M109, MSTA, Panzerhaubitze 2000), and support vehicles (military truck). Each class contains approximately 50 images. As most samples from this dataset were acquired from public sources, most images have a high resolution and contain centered objects. More information on the dataset can be found in van Woerden et al. (2025) \cite{van2025occlusion}. 

\textbf{2B: Ukraine military vehicles classification dataset} is an in-house dataset acquired from frames from three publicly available videos from Ukraine\footnote{\url{https://www.youtube.com/watch?v=3N6JKiYuo5w}; \url{https://www.youtube.com/watch?v=2TkxrxVlHyo}; \url{https://www.youtube.com/watch?v=Y9pLJXhFBqU}}. The 289 samples in this dataset are cropped bounding boxes containing vehicles present in the frames, which were manually labelled using function classes, e.g., armored personal carriers, military trucks and self-propelled artillery. As the samples are cropped from video frames, this dataset contains low resolution and challenging operational conditions.

\textbf{2C: STORE military object classification dataset} \cite{langlois2026} is a military dataset that was acquired from videos recorded during two campaigns, (1) at the military site in Fontevraud-l'Abbaye, France and (2) at a testing facility near Kongsberg, Norway; as part of the EDF project STORE. The objects present during these campaigns include military and civilian vehicles. The dataset was recorded using a mix of drones, stationary, and moving ground vehicles, hence includes videos from both an air-to-ground and ground-to-ground perspective. The dataset was recorded with the goal of representing an operational, military context and varying conditions: one portion was recorded in Nordic, snowy conditions and the other portion in a milder Mid-European climate with some rainfall. Objects could be occluded by trees or rainfall. The selected dataset contains 1,813 cropped bounding boxes obtained from 80 videos.

Besides two different domains with varying coverage in the VLM training data, the four datasets include different levels of challenging operational conditions. While dataset 2A consists of military vehicles, most images are high-quality and contain centered objects, making it more aligned with typical academic settings. In contrast, dataset 2B contains images from an operational environment and 2C contains images from a campaign meant to simulate operational conditions, including adverse weather conditions, camouflage and small objects. The images in 2C are also not publicly available, meaning that the VLMs could not have been trained on them, whereas it is possible that the images in 1A, 2A and 2B were part of the training data of the VLMs that we used. Additionally, datasets 2B and 2C were obtained by cropping frames in videos, resulting in lower quality and partially occluded objects. Moreover, as these datasets were acquired from video frames, they have a smaller variation in surroundings compared to dataset 2A. We hypothesize that more variation in the environment will benefit the class-level average embeddings $e_c$ and, in turn, the quality of the derived clusters. Therefore, this work includes experiments on a single military vehicle dataset, as well as a combination of the three military vehicle datasets.

\begin{figure}[ht!]
    \centering
    \includegraphics[width=\textwidth]{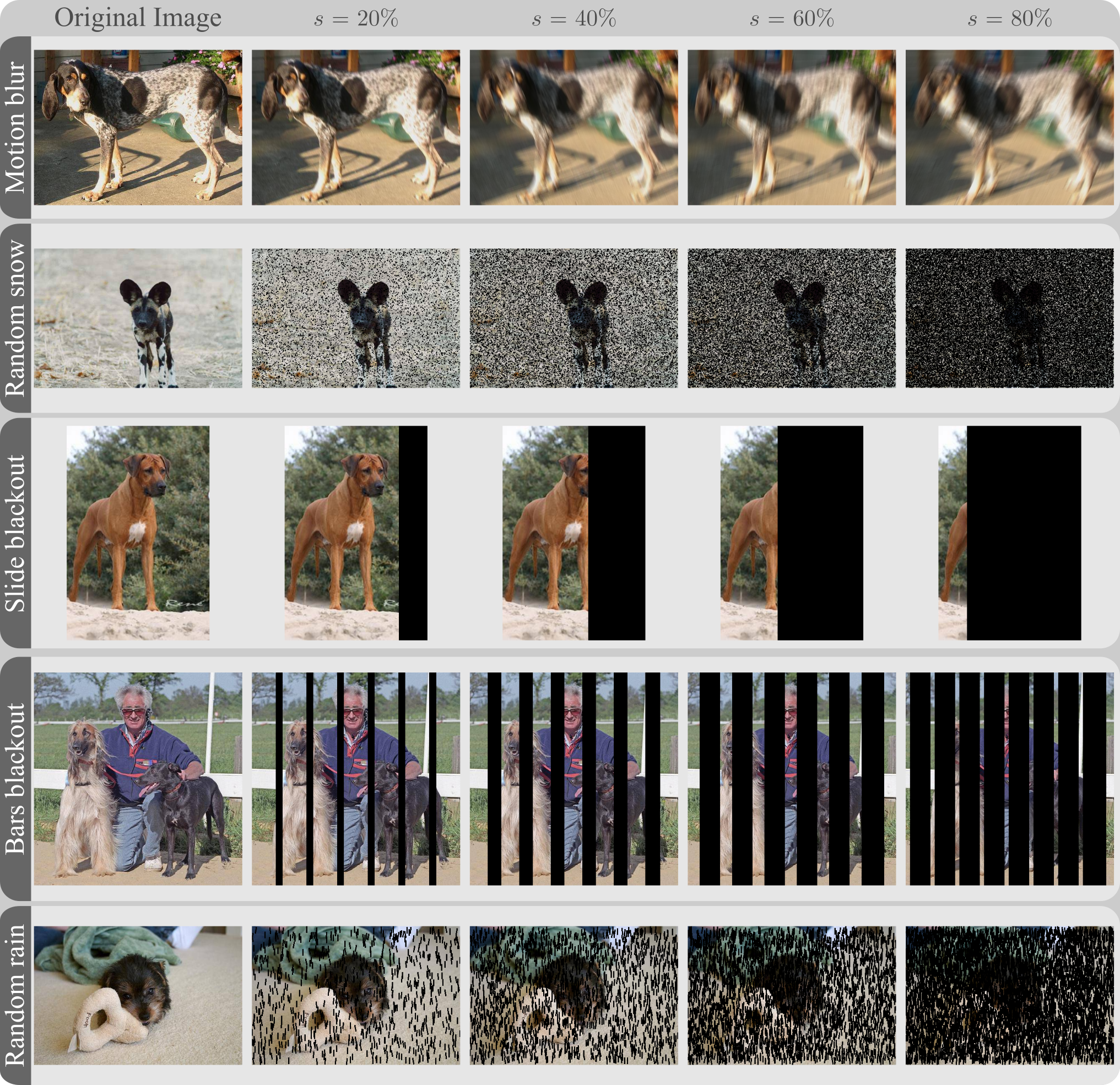}
    \caption{Examples of the different perturbations and occlusions that are used to corrupt our data, for increasing severities $s \in \{20, 40, 60, 80\}$ \%.}
    \label{fig:corruptions}
\end{figure}

\subsubsection{Introducing perturbations}
We assess how well our VLM-based ESD method captures and clusters systematic challenges in the data. We add artificial perturbations to the image data to simulate operational challenges \cite{van2025occlusion}. Motion blur, random snow, slice blackout, bars blackout and random rain are applied at various severities: $s \in \{20 \%, 40 \%, 60 \%, 80 \%\}$, as shown in \autoref{fig:corruptions}.

\subsection{Evaluation}
We evaluate the derived clusters both qualitatively and quantitatively. For quantitative evaluation, we use the silhouette score \cite{rousseeuw1987silhouettes} of the derived clusters in the embedding space. This score measures how similar each sample is to its own cluster compared to the other clusters and ranges between -1 and 1, where higher is better. We qualitatively evaluate the cluster descriptions obtained from the generative VLM and the semantic content from the images in the cluster. Here we focused on overlap and distinctiveness of the cluster descriptions. Moreover, we manually assessed correctness for the top descriptors in dataset 1A.

\subsection{Implementation details}
This section provides the implementation details of the models, algorithms and datasets used in our experiments.

\subsubsection{Classification models}
Since we are studying ESD methods with the purpose of V\&V, we introduce several classification models as \textit{models under evaluation}. Note that since our focus is on the identification of systematic errors, training a perfect classification model is outside the scope of this work. We used different models for dog breed and military vehicle classification. For dog breed classification, we used YOLO26-cls, version L \cite{sapkota2025yolo26}. While limited resources were spent on fine-tuning this model for dog-breed classification (50 epochs), it achieved an accuracy of 88\% on the validation set and 90\% on the held-out test set. For military vehicle classification, we leveraged CLIP PE-Core-L-14-336 \cite{cho2026perceptionlm,bolya2026perception} as a zero-shot classifier based on initial experiments showing its strong baseline performance for the military domain. Evaluation was conducted only on the Ukraine military vehicle classification dataset (2B \autoref{fig:dataset}) and resulted in varying performance depending on the video quality. Videos 1 and 3 resulted in an accuracy of 81\% and 88\%, respectively, while the classification accuracy in video 2 was only 45\%.

\subsubsection{VLMs}
We made use of two different VLMs for steps \Circle{2}, \Circle{3} and \Circle{5} of our method (\autoref{fig:method}). For obtaining the embeddings of the images in the joint latent space, we used the contrastive CLIP PE-Core-L-14-336 \cite{cho2026perceptionlm,bolya2026perception} model. This model was also used to obtain the class-level average embeddings $e_c$ for the available classes in the dataset. For obtaining the cluster descriptions, we used the generative GPT-5 mini model \cite{singh2025openai}, as this provided the best trade-off between high performance and computational efficiency. Given a collage of 10 images for each cluster, we prompted this model to give a description of the top-5 commonalities between the images, ranked from most important to least important. The full prompt is given in \autoref{app:prompt}.

\subsubsection{Data splits} 
The dog breed classification dataset (1A, \autoref{fig:dataset}) was provided with a training, validation and test split consisting of 6391, 762, and 887 images, respectively. As this dataset contained an abundance of samples, we fine-tuned the classification model on the training split, and used the validation split for obtaining the class-level average embeddings ($e_c$). The test split was used to validate the full ESD pipeline, as described in \autoref{fig:method}. In contrast, the Ukraine vehicle classification dataset (2B) was much smaller and hence all images (289) were used both for obtaining class-level average embeddings and for validation the full ESD pipeline. Since a zero-shot classifier was used, no training set was necessary. For the STORE dataset (2C), the average class embeddings were calculated based on 382 samples, held out from the test set with 1431 samples. Finally, 260 samples from the public military dataset (2A) were used for calculating the average class embeddings and 477 different samples were used for testing the ESD pipeline. 

\subsubsection{Clustering}
Prior to clustering of the residual embeddings in the joint latent space $\mathcal{E}$, we relied on non-linear dimensionality reduction to 10 dimensions using the UMAP algorithm \cite{mcinnes2018umap}. We applied this algorithm with Euclidean distance, a minimum distance of 0.0 and 15 neighbours. Subsequently, we used the HDBSCAN algorithm to identify clusters in this lower dimensional embedding space \cite{campello2015hierarchical}. For this algorithm we used a minimal cluster size of 5 and a minimum sample size of 25. These algorithms and parameter settings were selected based on them yielding the highest silhouette score on our dog breed classification dataset compared to other parameters when doing a full parameter sweep and methods, namely T-SNE, UMAP, Sparse autoencoder, PCA and Kernel PCA and K-means, HDBSCAN, DBSCAN, Agglomerative and Spectral clustering.

\section{EXPERIMENTS \& RESULTS}
In this section, we provide experimental results of our ESD method on both dog breed classification and military vehicle classification. We start with results on the dog breed data, a domain which is well represented in VLM training datasets. Here, we assess the cluster quality and the ability of the ESD method to capture artificially introduced perturbations (\autoref{fig:corruptions}). Subsequently, we move towards the domain of military vehicle classification, which has a lower representation in the VLM training data compared to the dogs. We assess the effect of this in terms of the cluster quality, but also include an analysis at the embedding level. We also include a comparative analysis between the two domains at the embedding level.

\subsection{Dog breed classification}
For our dog breed classification dataset, we assess if our ESD method can derive semantically meaningful clusters from a domain that is well represented in the VLM training data and contains largely varying environments. Moreover, we assess if our method can identify artificially introduced systematic perturbations in the data.

\subsubsection{Cluster quality}
\begin{figure}
    \centering
    \includegraphics{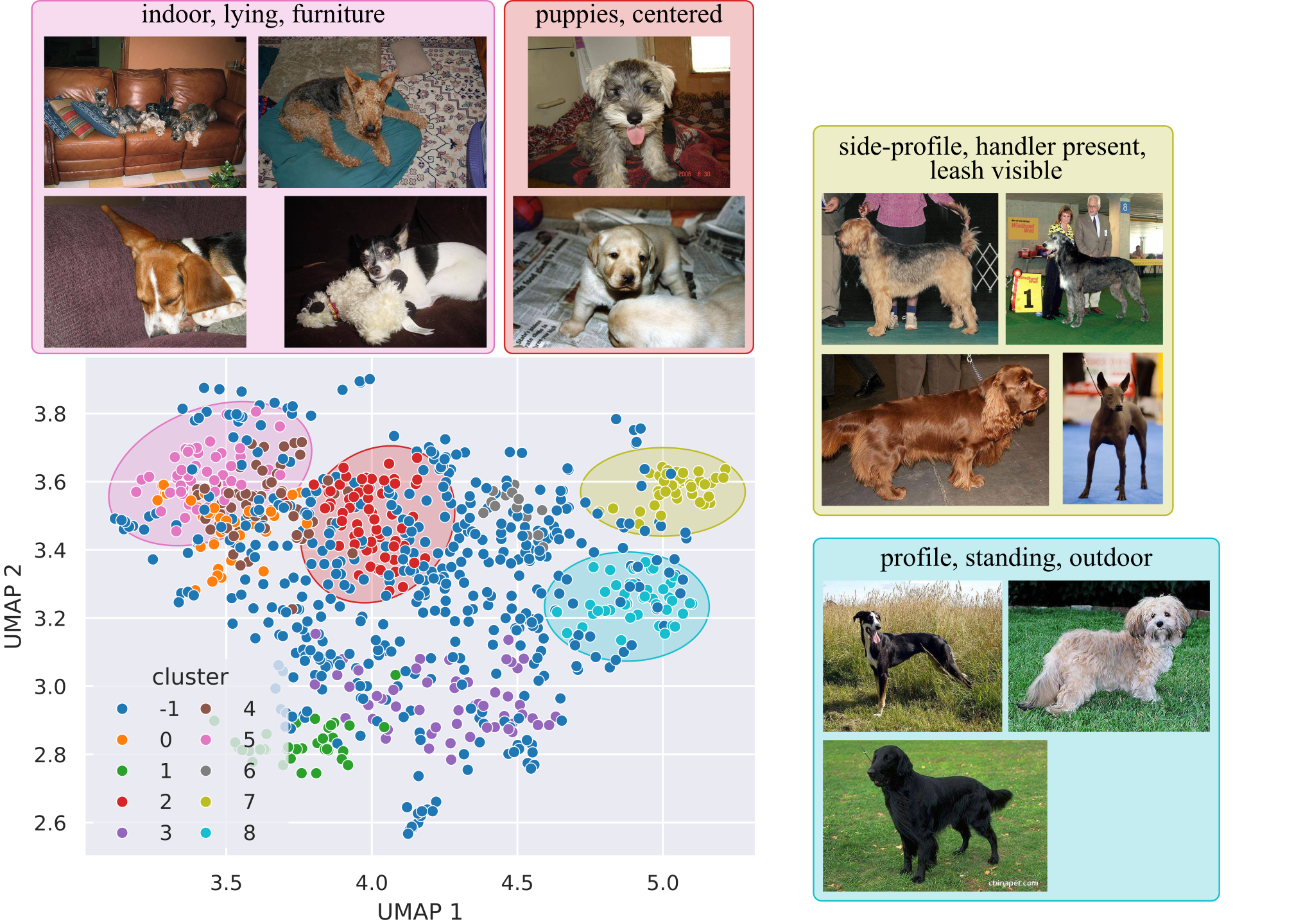}
    \caption{UMAP visualization of the residual embeddings of the dog breed dataset. Some representative examples, including cluster descriptors, are shown for four clusters, namely 2 (red), 5 (pink), 7 (yellow) and 8 (blue). Unassigned samples are indicated by cluster `-1'. Quantitative results for all nine clusters are shown in \autoref{tab:clusters_db_sst}.}
    \label{fig:dog_clustering}
\end{figure}

Our method yielded nine clusters on the dog breed dataset, achieving a silhouete score of 0.67. Moreover, the majority of image samples were not assigned to any cluster (469 images, 53\%). Nevertheless, the derived nine clusters, shown in \autoref{fig:dog_clustering}, display semantic coherence. For example, cluster 2 consists of puppies, cluster 5 contains dogs lying on furniture, and cluster 7 contains dogs competing in a dog show. \autoref{tab:clusters_db_sst} shows the VLM-derived cluster descriptions, which are largely in agreement with the depictions provided by the qualitative overview in \autoref{fig:dog_clustering}. To support this quantitatively, we manually assessed the presence of some of the attributes in the description and reported the accuracy. In the best case, 100\% of the images agreed with the attribute while this was 79\% for the worst performing cluster.

Moreover, we assessed the performance of the classification model on each of the clusters, which is included in \autoref{tab:clusters_db_sst}. This reveals that the classification performance on cluster 6 is far below the reported average performance of 90\%. Based on the description generated for this cluster, the model struggles with samples where the subject is centered in a cluttered background and subject to low resolution, blur and varied lighting conditions. In contrast, the model performs better on classifying dogs in the snow (cluster 1), or in the grass (cluster 3). 

\begin{table}[]
    \centering
        \caption{The clusters formed by the slice-then-tag method on the dog breed dataset, paired with the correctness of the first/most important word(s) (in bold), assessed through manual inspection, and the model accuracy for that cluster.}
    \label{tab:clusters_db_sst}
    \begin{tabular}{ccccc}
    \textbf{Cluster no.} & \textbf{\thead{Cluster description}} & \textbf{\thead{Images \\ assigned}} & \textbf{\thead{Cluster \\ correctness}} & \textbf{\thead{Model \\ accuracy}} \\
    \hline
    -1 & \makecell{side-facing, outdoor, low-angle, occluded, \\cluttered-background} & 469 & - & 91\% \\ \vspace{-1mm}
    0 & \makecell{\textbf{multiple}, occlusion, outdoor, side-view, lighting} & 37 & 100\% & 86\% \\ \vspace{-1mm}
    1 & \makecell{\textbf{snow}, overcast, centered, occluded, low-contrast} & 35 & 94\% & 94\%\\ \vspace{-1mm}
    2 & \makecell{puppies, centered, varied-backgrounds,\\similar-lighting, occlusion} & 77 & - & 87\%\\ \vspace{-1mm}
    3 & \makecell{\textbf{grass}, \textbf{outdoor}, centered, lighting, pose} & 53 & 81\%, 96\% & 96\%\\ \vspace{-1mm}
    4 & \makecell{\textbf{indoor}, centered, close-up, flash, top-down} & 65 & 86\% & 91\% \\ \vspace{-1mm}
    5 & \makecell{\textbf{indoor}, \textbf{lying}, closeup, furniture, dim} & 52 & 90\%, 87\% & 87\%\\ \vspace{-1mm}
    6 & \makecell{\textbf{centered subject}, similar poses, cluttered \\backgrounds, low resolution/blur, varied lighting} & 14 & 93\% & 71\%\\ \vspace{-1mm}
    7 & \makecell{\textbf{side-profile}, standing, handler-present,\\ leash-visible, cluttered-background} & 42 & 86\% & 93\%\\ \vspace{-1mm}
    8 & \makecell{\textbf{profile}, fur, standing, outdoor, occlusion} & 43 & 79\%  & 93\% 
    \end{tabular}
\end{table}

\newpage
\subsubsection{Systematic perturbations}
In this experiment, we simulate systematic challenges in the data by randomly applying five occlusions and perturbations shown in \autoref{fig:corruptions} to the images prior to acquiring VLM embeddings. For each sample in our test set, we applied one of the five transforms at a severity $s \in \{20\%, 40\%, 60\%, 80\% \}$ randomly sampled from a uniform distribution. Based on these perturbations, the performance of our classifier dropped from 90\% on the clean test set to 38\% on the perturbed set. Subsequently, we let our ESD method find clusters of related samples. This results in three clusters, achieving a silhouette score of 0.87. This indicates an improved coherence and separation of the clusters compared to the previous experiment. Moreover, all samples were assigned to one of the three clusters.

\begin{figure}
    \centering
\includegraphics{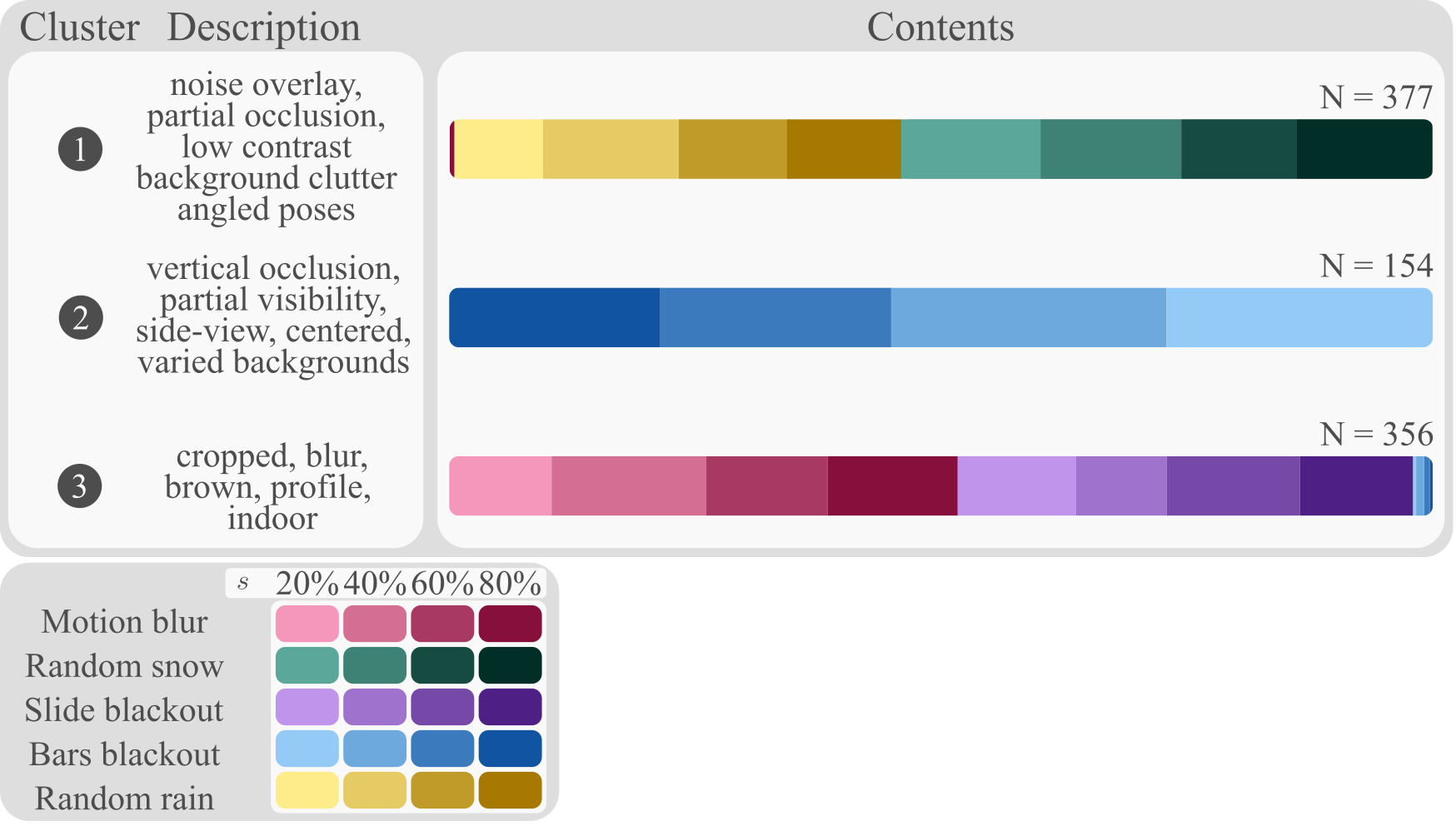}
    \caption{The number of images per cluster formed by the slice-then-tag method on the perturbed dog breed classification dataset, plotted against the added augmentations and their severity.}
    \label{fig:pre_dogs_results}
\end{figure}

\autoref{fig:pre_dogs_results} shows how the occlusions at different severities are distributed across the clusters. We observe that images transformed using \textit{random snow} and \textit{random rain} were placed together in cluster 1. Similarly, images that were perturbed using \textit{motion blur} and \textit{slide blackout} were placed in cluster 3, while images transformed using \textit{bars blackout} formed cluster 2 on their own. Only a small number of images that underwent a \textit{bars blackout} transform ended up in cluster 3. Note that the severity of the transform does not affect the assigned cluster, indicating that also subtle alterations of the data can be picked up by the ESD method. 

The VLM-derived cluster descriptions are also included in \autoref{fig:pre_dogs_results}. These seem to overlap with the factual description of the transformation applied to the images, e.g. `noise-overlay' and  `partial occlusion' describing the \textit{random rain} and \textit{random snow} perturbations and `vertical occlusion' describing \textit{bars blackout}. Cluster 3 contains both the \textit{motion blur} and \textit{slide blackout} transforms and is described as `cropped', potentially referring to the partial coverage of the image by the blackout area, and `blur', aligning with the \textit{motion blur} transform. Interestingly, the final descriptor `indoor' refers to an environmental factor rather than one of the perturbations, though, upon manual inspection, that environmental factor does not seem to be consistently present in the images assigned to that cluster.

We conclude that our ESD method is able to detect and categorize consistent challenges in the image data. Notably, many of the identified phenomena correspond to operational conditions rather than physical objects, suggesting that the VLM embedding space captures higher-level scene characteristics beyond object-level semantics.

\subsection{Military vehicle classification}

Based on these initial experiments, we next investigate how well our ESD method works in the military domain. We investigate if this method can also form semantically coherent clusters of military vehicle imagery.

\subsubsection{Cluster quality}

\begin{figure}
    \centering
    \includegraphics{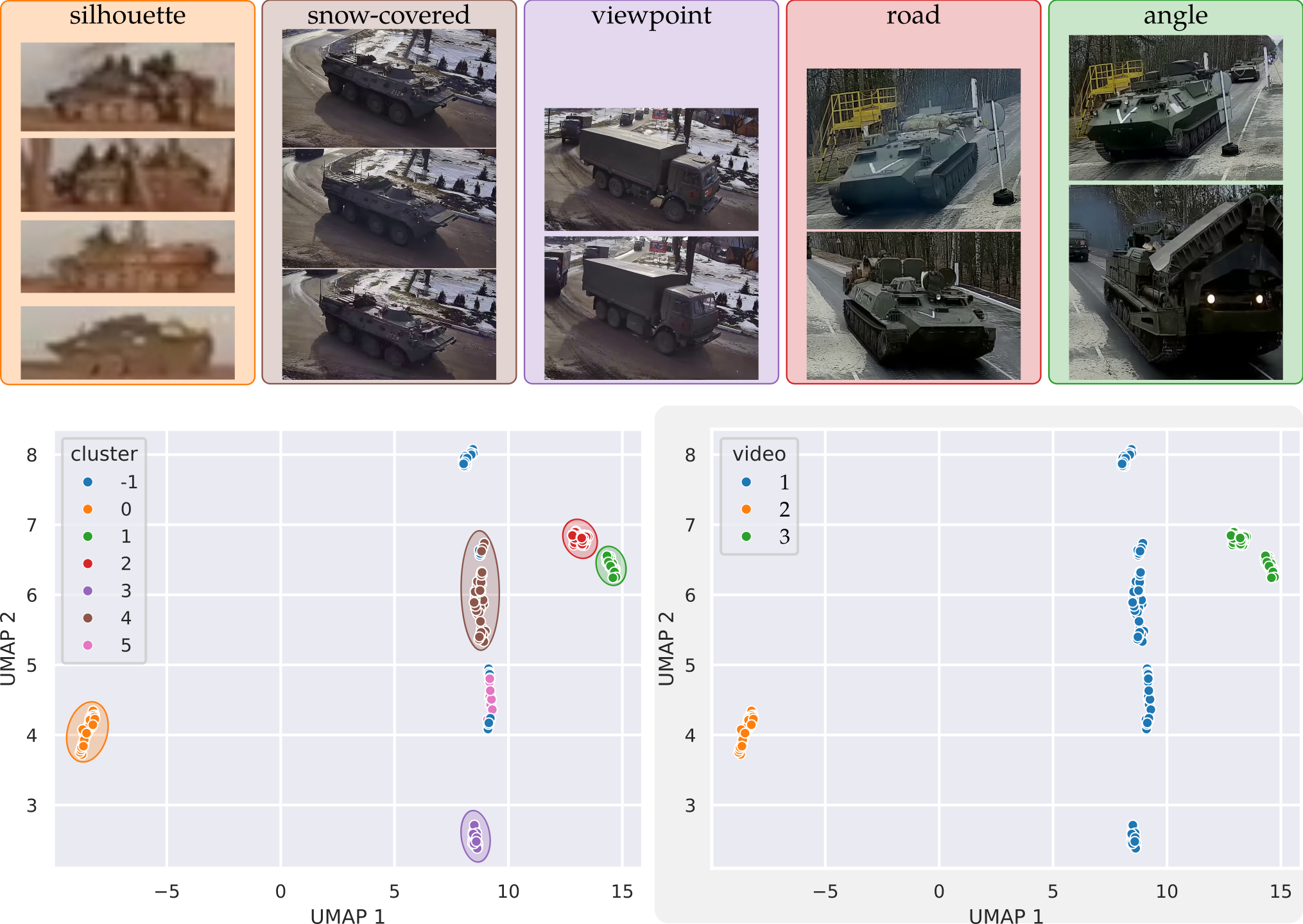}
    \caption{A visualization of the clusters detected on the Ukraine military vehicle classification dataset (2B) with the first word of the corresponding cluster description. The bottom right plot shows the same UMAP plot, but colored per video i.o. per cluster.}
    \label{fig:ukr_cluster}
\end{figure}

Applying the ESD method to the Ukraine vehicle classification dataset (2B) resulted in a relatively high silhouette score of 0.83 with 46 unassigned samples, indicating that the residual embeddings formed tight and well-separated clusters.  \autoref{fig:ukr_cluster} shows that the derived clusters are more disjoint than those of the dog breed dataset in \autoref{fig:dog_clustering}. However, the cluster descriptions in \autoref{tab:clusters_ukraine_stt} are less distinct and show considerable overlap, making them less informative than those obtained for the dog breed dataset. Closer inspection revealed that the derived clusters largely overlap with the three distinct videos that make up dataset 2B: cluster 0 corresponded nearly entirely with video 2, while videos 1 and 3 were distributed across multiple clusters (bottom right,  \autoref{fig:ukr_cluster}). This also explains recurring cluster descriptions, such as snow-related terms for video 1. Moreover, the description for cluster 0, containing low-quality footage from video 2, does not mention this obvious aspect in the description. In addition, it seems that the vehicle class is not fully separated from the derived clusters. For example, \autoref{fig:ukr_cluster} shows images from video 1 in clusters 3 and 4, each containing different vehicles as the only obvious differentiating characteristic. In conclusion, the derived clusters and descriptions have limited additional value in this setting, as the same information could have been derived from existing metadata.

\begin{table}[p]
    \centering
    \caption{The clusters formed by the slice-then-tag method on dataset 2B, paired with the number of images assigned to that cluster and the model accuracy for that cluster.}
    \label{tab:clusters_ukraine_stt}
    \begin{tabular}{cccc}
    \textbf{ Cluster no.} & \textbf{\thead{Cluster description}} &\textbf{\thead{Images\\ assigned}}  & \textbf{\thead{Model\\ accuracy}} \\ 
    \hline
     -1 & \makecell{snowy-road, overhead-angle, low-contrast,\\ large-dark-silhouettes, partial-occlusion} & 46 & 89\% \\ \vspace{-1mm}
     0  & \makecell{silhouette, color, angle, background, lighting} & 57 & 44\% \\ \vspace{-1mm}
     1 & \makecell{angle, lighting, color, background, occlusion} & 34 & 82\%\\ \vspace{-1mm}
     2 & \makecell{road, forest, frontal, overcast, occluded} & 50 & 84\%\\ \vspace{-1mm}
     3 & \makecell{viewpoint, background, lighting, occlusion, snow} & 26 & 100\%\\ \vspace{-1mm}
     4 & \makecell{snow-covered background, armored vehicles,side view, \\ low-contrast lighting, partial occlusions} & 61 & 79\%\\ \vspace{-1mm}
     5 & \makecell{viewpoint, snow, low-contrast, motion-blur, occlusion} & 15 & 100\%
    \end{tabular}
\end{table}

\begin{figure}
    \centering
    \includegraphics[width=0.9\textwidth]{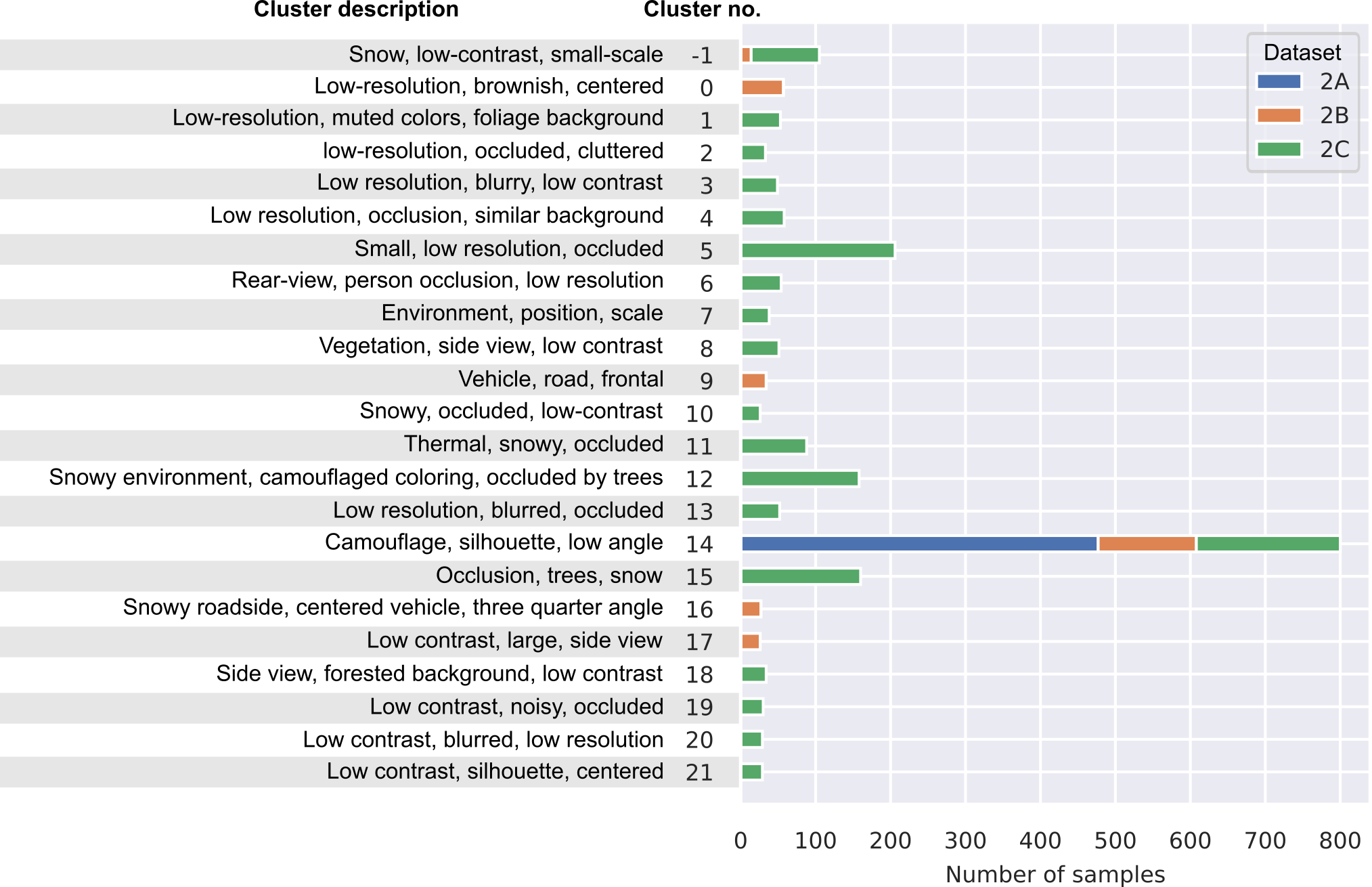}
    \caption{Overview of the clusters derived from a combination of dataset 2A, 2B and 2C. Each row gives the distribution of the cluster across the three datasets, as well as the top-3 automatically generated descriptions.}
    \label{fig:dist_allmil_clusters}
\end{figure}

\begin{figure}
    \centering
    \includegraphics[width=0.85\textwidth]{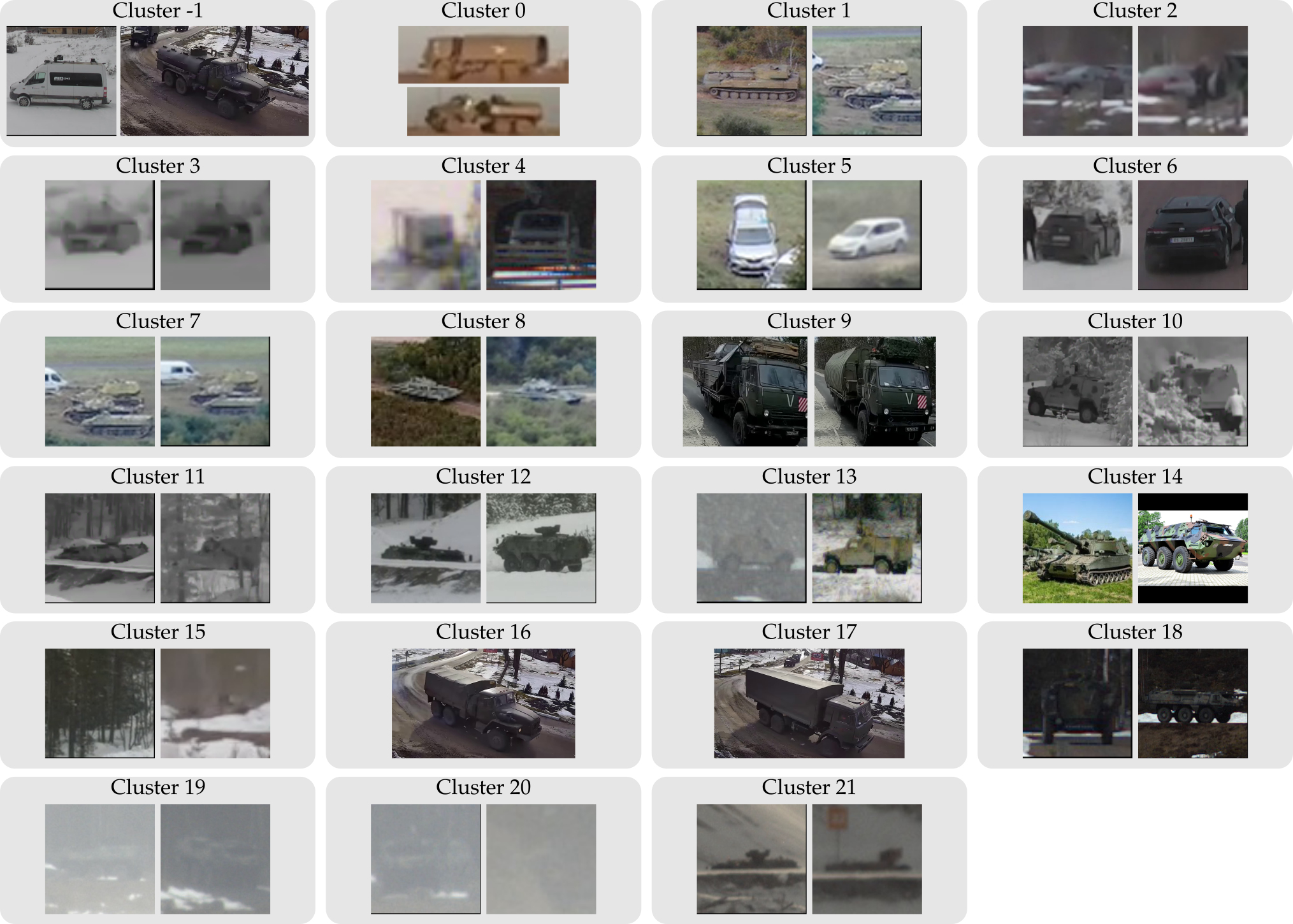}
    \caption{Visualization of the clusters derived from the combination of datasets 2A, 2B and 2C.}
    \label{fig:vis_all_mil_clusters}
\end{figure}

The previous experiment demonstrated that the clustering was dominated by video-specific characteristics, resulting in limited new insights. To evaluate the utility of our ESD method in a broader military context, we combined datasets 2A, 2B and 2C prior to embedding and clustering. This experiment included 2,197 images of military vehicles. The method yielded 22 different clusters of varying size (\autoref{fig:dist_allmil_clusters}). Analysis of the dataset distribution across clusters revealed that most clusters contained samples from only a single dataset, indicating that clustering was largely driven by dataset-specific characteristics. Cluster 14 was the only notable exception and was substantially larger than the other clusters, containing all samples from dataset 2A as well as samples from 2B and 2C. \autoref{fig:dist_allmil_clusters} also presents the top 3 of the cluster descriptions. Similar to the labels provided in \autoref{tab:clusters_ukraine_stt}, these descriptions contain recurring elements (e.g. `low resolution', `small', `low contrast', `occluded'), limiting their value for differentiating between clusters. This lack of distinctiveness is further reflected by the cluster visualizations shown in \autoref{fig:vis_all_mil_clusters}, revealing that the differences between clusters are often subtle. 

\subsection{Dataset analysis}
\label{sec:datasetanalysis}
We observed a performance discrepancy for our ESD method between the dog breed classification and the military vehicle classification datasets. We hypothesize this is related to a lack of variation in the environment of the main object and evaluate if this can be derived by comparing the variation in the (residual) image embeddings for both datasets.

\begin{figure}
    \centering
    \includegraphics[width=0.985\textwidth]{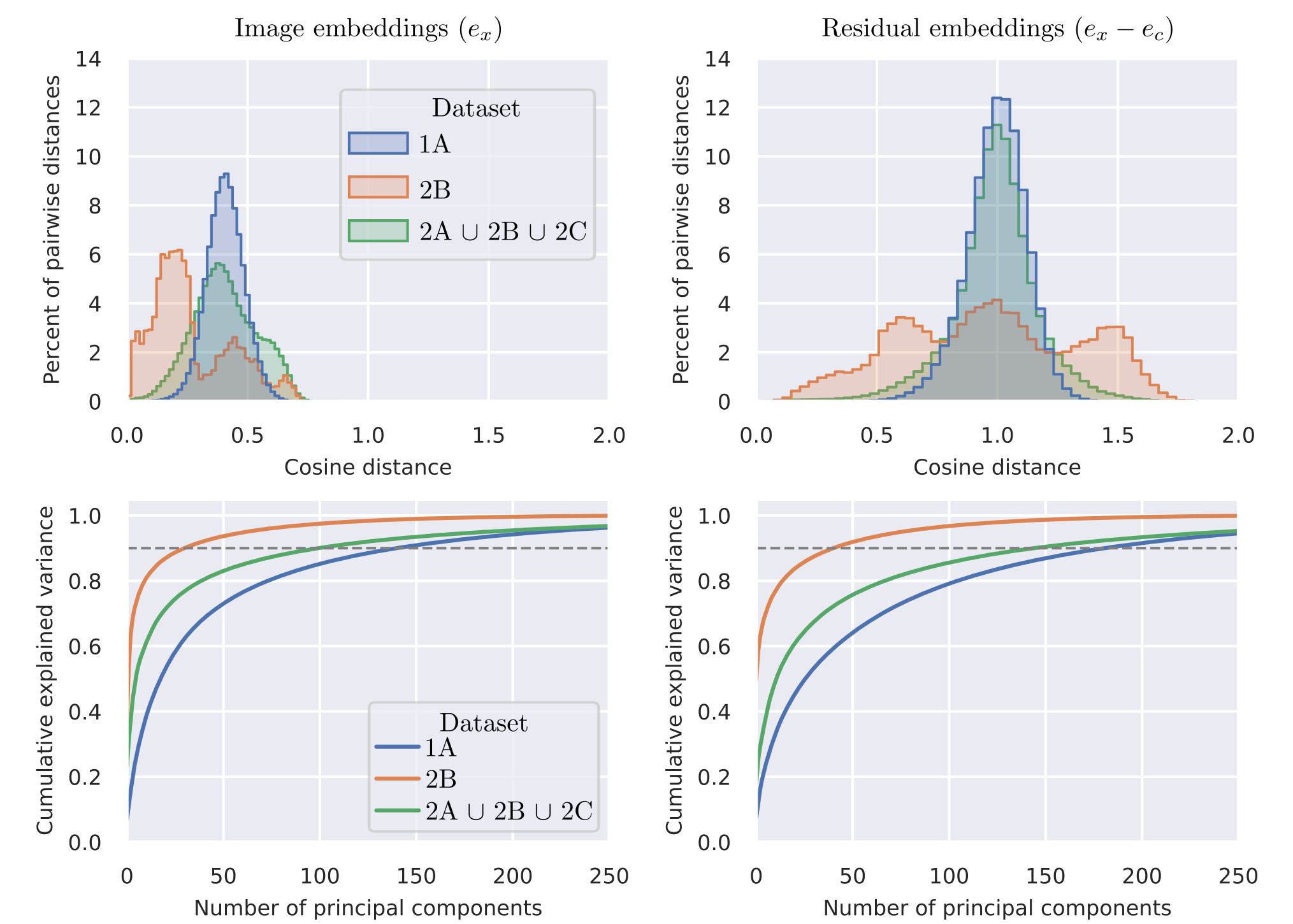}
    \caption{Comparison between the dog breed dataset and the military vehicle datasets. Top row: distribution of pairwise cosine similarities for the image embeddings (left) and the residual embeddings (right). Bottom row: cumulative explained variance from the PCA analysis for the image embeddings (left) and the residual embeddings (right).}
    \label{fig:dataset_analysis}
\end{figure}

We measure the variation of both the raw image embeddings and the residual embeddings using the distribution of pairwise cosine distances and the cumulative explained variance from PCA analysis. \autoref{fig:dataset_analysis} shows the results of this comparison for dataset 1A, 2B and a combination of 2A, 2B and 2C. The distribution of pairwise cosine distances of the image embeddings of dataset 1A roughly follows a normal distribution. In contrast, dataset 2B clearly shows three local maxima, representing the three videos this dataset consists of. The distribution of cosine distances for the combination of the three military vehicle datasets is more similar to the dog breed dataset, but has one additional local maximum. This local maximum disappears when the pairwise distances between residual embeddings are considered. Nevertheless, the local maxima in the distribution of dataset 2B persists. Moreover, the cosine distances between samples increase for all datasets after the mean class embedding is subtracted. This implies a larger variance in all residual embeddings compared to the original embeddings. Still, a clear difference in variance between the dog breed dataset and the combination of the three military vehicle datasets is not observed from these plots.

The PCA analysis in the bottom row of \autoref{fig:dataset_analysis} tells a similar story as the cosine distances in the top row. Here, we see that more than 90\% of the variation in dataset 2B is explained in the 30 first principle components. Meanwhile, this amount of variance is explained in the first 99 and 140 components for all military vehicle datasets and the dog breeds dataset, respectively. This implies a much more limited variation in the embeddings of dataset 2B and a higher variation for the datasets 2A and 2C. The residual embeddings require more principal components to explain 90\% of the variation, which implies that the residual embeddings show more variation than the original embeddings.

\begin{figure}[t!]
    \centering
    \includegraphics[width=0.6\textwidth]{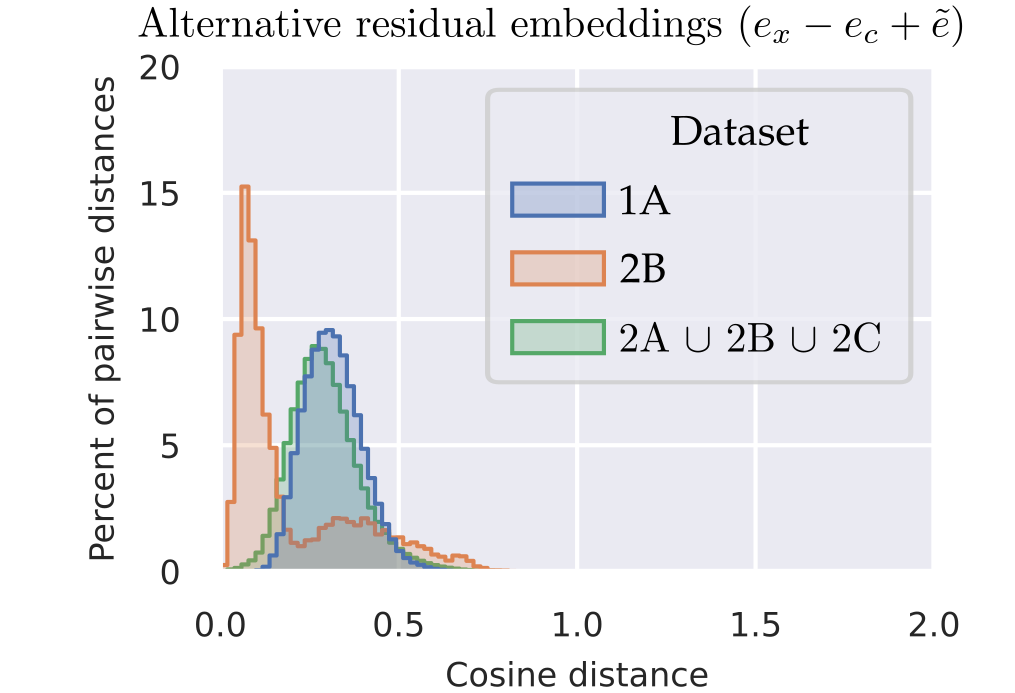}
    \caption{Distribution of cosine similarities for the three data cohorts for the residual embeddings corrected with the dataset level average embeddings.}
    \label{fig:alternative_embeddings}
\end{figure}

The increased cosine similarities and variation after subtracting the class mean embeddings suggest that a dominant component is removed and individual variations are amplified, potentially at the expense of dataset-level semantics. As the cosine similarity is based on directions, subtracting the mean class-wise direction may distort the directions in the resulting residual embeddings. We investigate this effect by recomputing the pairwise distances after adding the \textit{dataset}-level average embedding ($\tilde{e}$) to the residual embeddings. \autoref{fig:alternative_embeddings} shows that this consistently reduces the pairwise distance compared to the original embeddings, as shown in the top left of \autoref{fig:dataset_analysis}. This could imply that adding the dataset-level average embedding removes class-level information and preserves dataset-level information, while this information may be lost in embeddings where only the class-level average is subtracted.

\subsection{Residual embedding}
\label{sec:residualembedding}
Subtracting the mean class embedding is an important step in the proposed ESD method. In this section, we assess the effect of this operation separately. We qualitatively compare the image embeddings before and after subtracting the mean class embedding for different dataset compositions. In addition, we compute the silhouette score using the class descriptions as cluster assignments to quantitatively assess the class-related structure in the embeddings.

\begin{figure}
    \centering
    \includegraphics[width=0.9\textwidth]{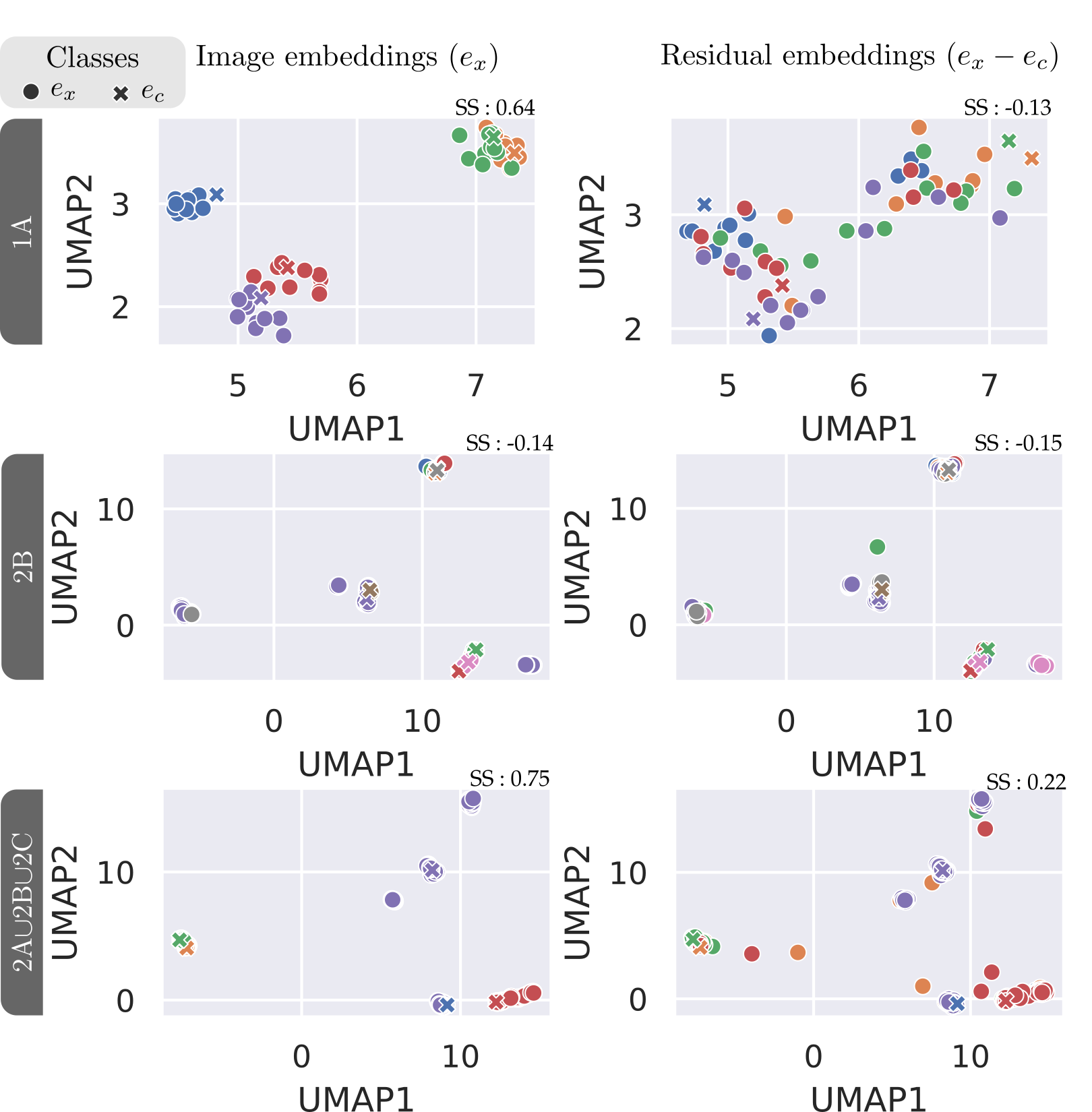}
    \caption{UMAP visualizations of the original image embeddings (left column) and the residual embeddings (right column) for the dog breed dataset (top row) Ukraine military vehicle classification datset (middle row) and the combination of all three military vehicle classification datasets (bottom row). Different classes are indicated by marker colors, embeddings from images ($e_x$) are indicated as \faCircle, while mean class embeddings ($e_c$) are indicated as \faRemove. The top right of each plot shows the silhouette scores calculated from the class descriptions.}
    \label{fig:res_emb}
\end{figure}

\autoref{fig:res_emb} shows that the dog breeds form a dominant component in the image embeddings of dataset 1A, as they are strongly organised according to their class descriptions. This observation is reflected by the relatively high silhouette score of 0.64. After subtracting the mean class embeddings, the classes become substantially more mixed, resulting in a silhouette score of -0.13. The UMAP embeddings were based on the image embeddings ($e_x$) and separately determined for each dataset cohort in \autoref{fig:res_emb}. We used the same UMAP projection for the image and residual embeddings, enabling a direct visual comparison of the resulting structures. These results suggest that subtracting the mean class embeddings removes a dominant class-related feature from the embeddings, allowing other sources of variation in the image data to become more prominent. 

In contrast, for dataset 2B, embeddings of images that belong to the same class are not consistently grouped together. This indicates that the object class constitutes a less dominant component of the embedding representation. Consequently, subtracting the mean class embeddings has a smaller effect on the overall embedding structure, which is reflected by the nearly unchanged silhouette score before and after subtraction. The combination of datasets 2A, 2B and 2C shows an intermediate effect. Here, the image embeddings are organized by class, resulting in a high silhouette score of 0.75. Subtracting the mean class embedding reveals a different structure, where embeddings of classes are more mixed. Likewise, the silhouette score decreases to 0.22, indicating that this effect is less pronounced than is the case for dataset 1A.

\newpage
\section{DISCUSSION}

In this study, we have presented several experiments using a VLM-based ESD method to automatically discover systematic errors in an AI model. We have focused on the ability of this method to detect systematic errors in dog breed and military vehicle classification datasets. Based on the results, we have further investigated the differences in variation of embeddings and residual embeddings for aforementioned datasets. 

Our method showed promising results on the dog breed classification dataset. In particular, it successfully identified the systematic vulnerabilities that were deliberately introduced through image perturbations. Furthermore, subtracting the mean class embedding resulted in meaningful clusters corresponding to different environments. \autoref{fig:res_emb} confirms that the class forms a prominent part of the embeddings for dogs, and that subtracting the mean class embedding reveals novel structures in the embedding space. These findings suggest that the approach can detect and categorize consistent challenges and higher-level scene characteristics in the image data. However, a notable limitation is that the majority of samples were not assigned to any cluster, reducing the overall coverage of the analysis. This may be a consequence of the clustering algorithm used, and alternative clustering approaches could potentially improve the fraction of data captured in meaningful slices.

In contrast to the dog breed dataset, our method yielded fewer novel insights into systematic vulnerabilities in the military vehicle datasets. Discovered cluster descriptions overlapped and clusters corresponded to individual videos, though this was expected given the goal of our method. Videos naturally exhibit the same environmental conditions, hence deriving clusters that match the different videos is in line with the results on the dog breed dataset. We expect that this method is most valuable for datasets that are derived from separate images, which contain more varying environmental conditions. However, we also suspect that subtracting video average embeddings rather than class average embeddings may be a promising route towards better results for video data.

A prerequisite for the residual embeddings leveraged by our ESD method is \textit{concept purity} of the VLM in the domain it was used on. This entails that image embeddings are composed of a linear combination of the main object and its surroundings, i.e. $e_x \approx e_c + e_{\text{environment}}$. Moreover, in order to obtain accurate mean class embeddings, the object should be in different environments, to average out the environmental component in the image embeddings. We observed two important aspects that differ between the dogs and military vehicle images that may hinder the effectiveness of residual embeddings for the military dataset. First, military vehicles are underrepresented in the training data compared to different dog breeds, as these are more widely available on the internet. This affects the quality of the learned class representation. Second, the surroundings of the military vehicles are monotonous compared to the dog breeds dataset. The embeddings of military vehicles may contain parts of the environment as a result of these spurious correlations \cite{sagawa2019distributionally}. \autoref{fig:res_emb} supports this claim; for the different dog breeds, the image embeddings are mostly focused on the specific class, possibly because during training these dog breeds were seen in different environments. In contrast, the embeddings of military vehicles are not as focused on the specific class. Therefore, subtracting the mean class embedding for military vehicles results in a less pronounced structural change compared to the dog breeds. This effect may be mitigated by adding the dataset-level average embedding to the residual embeddings, as supported by the results in \autoref{fig:alternative_embeddings}. A reduction in the pairwise distances suggests the class average embedding contributes substantially to the total variation. Correcting the residual embeddings with the dataset average reduces the variation in the dataset. Future work should investigate if this results in more meaningful clusters, particularly for data-scarce domains.

\subsection{Limitations}

Several limitations of this study should be taken into consideration. First of all, the availability of public military vehicle classification datasets is overall limited. This study included 3 such sets, of which one was recorded in an actual operational environment (2B) and one in an environment meant to simulate operational conditions (2C). Both had shortcomings: 2B only included images from 3 videos and 2C was also recorded across only two locations. This limited environmental variability for both datasets could cause the clustering to stir towards videos/environments and thus limit the ability of our method to form clusters beyond those characteristics, as discussed in the previous paragraph.  

Another shortcoming of this study concerns the method of evaluation, especially quantitative evaluation. The V\&V process inherently relies strongly on qualitative assessment and though this study attempted to reduce this reliance, it in itself required qualitative evaluation. The silhouette score was used as an indicator of the cohesion of the clusters formed but it is not a representative metric of the usefulness of those clusters. Therefore, qualitative assessment had to be included, though it is subject to subjectivity and bias of the authors. 

Lastly, \autoref{sec:datasetanalysis} and \autoref{sec:residualembedding} hypothesize on the applicability of VLMs to the military domain in general and the usefulness of the residual embedding. Though these sections offer first insights into these domains, they do not offer a full ablation study of the different aspects impacting the performance of our method, namely the VLM selected, the knowledgeability of the VLM in the given domain, the calculation of the average class embedding and the residual embedding.

\subsection{Future work}

We identify several courses for future research. First of all, future work should address concept purity of VLMs in the military domain. The lack of concept purity and the existence of spurious correlations seemed to have limited the effectiveness of the use of residual embeddings in this domain. We identified two sources for this effect: underrepresentation in the training data, and limited variation of surroundings. Our experimental results are not conclusive for the individual contributions of both aspects and further investigation is required to understand the impact of and connection between these aspects. Moreover, the use of residual embeddings could be extended to an iterative method, where cluster average embeddings are iteratively subtracted to progressively reveal more subtle factors of variation and potential error modes.

Additionally, the current work attempted to derive cluster labels based on feeding a sample of images with a prompt to a generative VLM instead of using the embeddings that actually informed the formation of those clusters. This approach required using a different VLM for the cluster descriptions than for deriving the clusters, potentially affecting the distillation of the underlying semantic concepts found in the embedding space. Other methods for deriving the cluster descriptions should be considered in future work. For example, dictionary learning could be explored \cite{bricken2023towards}. This would allow us to make the transfer from embedding space to semantic space instead, potentially leading to results that allow for more differentiating cluster labels.

This study set out to accelerate the V\&V process of black-box third party AI systems. This acceleration can be promoted not only through full automation but also through supporting, prioritizing and steering a human analyst with VLM-based methods. Besides the technical improvements of the method and underlying principles, future work should therefore also focus on developing AI-assisted workflows that combine the scalability of VLMs with human expertise and judgment.

\section{Conclusion}

We conclude that current VLM-based ESD methods are not yet suitable to fully automate the V\&V process of AI models. However, they do show potential to accelerate the V\&V process by helping human analysts identify and characterize potential failure modes, especially in third party models. Additionally, the semantically structured embedding space provided by VLMs offers opportunities to characterize training and test data, identify underrepresented scenarios and reveal coverage gaps. However, the effectiveness of these approaches depends on both the target domain and the level of semantic interpretation required. Rather than fully automating V\&V, our findings suggest that the most promising role of VLMs is to support human analysts in discovering, prioritizing, and investigating systematic model errors. 

\acknowledgments 

The authors used ChatGPT (OpenAI) for language and grammar refinement of the manuscript text and the generation of code for conducting experiments; all scientific content and interpretations are the sole responsibility of the authors.

This work received funding from the European Defence Fund through the project STORE (Shared daTabase for Optronics image Recognition and Evaluation), grant agreement №101121405. We would like to particularly express our gratitude to all consortium partners involved in data acquisition and processing.

\bibliographystyle{spiebib} 
\bibliography{report}

\appendix
\section{prompt for cluster description generation}
\label{app:prompt}
The full prompt fed into GPT-5 mini, paired with 10 sampled images:  
``Please describe the similarities between these images, focussing on the factors that might make a classification model misclassify them, for example the environment (e.g. forest, snow, urban, road), position of the main object, lighting conditions, backgrounds, angle, or any image perturbations, augmentations or object occlusions. Please do not focus on the [dog breed/class]. Please respond with maximum 5 words, separated by commas, starting with the most important similarity." 

\end{document}